\documentclass{article}

\usepackage{microtype}
\usepackage{graphicx}
\usepackage{subfigure}
\usepackage{booktabs} 
\usepackage{array, multirow}
\usepackage{hyperref}
\usepackage{amsmath}

\usepackage[accepted]{icml2019}

\icmltitlerunning{Eigenvalue Spectra Analysis of Recurrent Models}

\begin{document}

\twocolumn[
\icmltitle{Evaluating the Stability of Recurrent Neural Models during Training with Eigenvalue Spectra Analysis}




\begin{icmlauthorlist}
\icmlauthor{Priyadarshini Panda}{to}
\icmlauthor{Efstathia Soufleri}{to}
\icmlauthor{Kaushik Roy}{to}

\end{icmlauthorlist}

\icmlaffiliation{to}{Department of Electrical \& Computer Engineering, Purdue University, West Lafayette, USA}

\icmlcorrespondingauthor{P. Panda}{pandap@purdue.edu}

\icmlkeywords{Reservoir Computing, Eigenvalue Spectra, Stability Analysis}

\vskip 0.3in
]



\printAffiliationsAndNotice{}  

\begin{abstract}
We analyze the stability of recurrent networks, specifically, reservoir computing models during training by evaluating the eigenvalue spectra of the reservoir dynamics. To circumvent the instability arising in examining a closed loop reservoir system with feedback, we propose to break the closed loop system. Essentially, we unroll the reservoir dynamics over time while incorporating the feedback effects that preserve the overall temporal integrity of the system. We evaluate our methodology for fixed point and time varying targets with least squares regression and FORCE training \cite{sussillo2009generating}, respectively. Our analysis establishes eigenvalue spectra (which is, shrinking of spectral circle as training progresses)  as a valid and effective metric to gauge the convergence of training as well as the convergence of the chaotic activity of the reservoir toward stable states. 
\end{abstract}

\section{Introduction}
Recurrent neural networks, specifically reservoir computing models, are studied in the context of neuroscience and neuromorphic computing to model and process inputs with spatio-temporal dynamics. A reservoir model is a randomly connected system of neurons that creates a complex, high dimensional dynamic representation of an input. The patterns of activity generated by the reservoir are then processed by a layer of linear readout neurons to perform pattern recognition tasks (Fig. 1). The intrinsic recurrence of such systems gives them a sort of `memory' to store patterns of correlated activity in a sequential input. Reservoirs also exhibit `chaos' on account of the recurrence, that presents a challenge in theoretical understanding and exploiting the nonlinear dynamics of such networks. However, understanding the dynamics of these networks during training has large implications toward advancing the reservoir computing field for broader range of artificial intelligence and computational neuroscience applications. 

\begin{figure}[!t]
\centering
\includegraphics[width=0.5\textwidth]{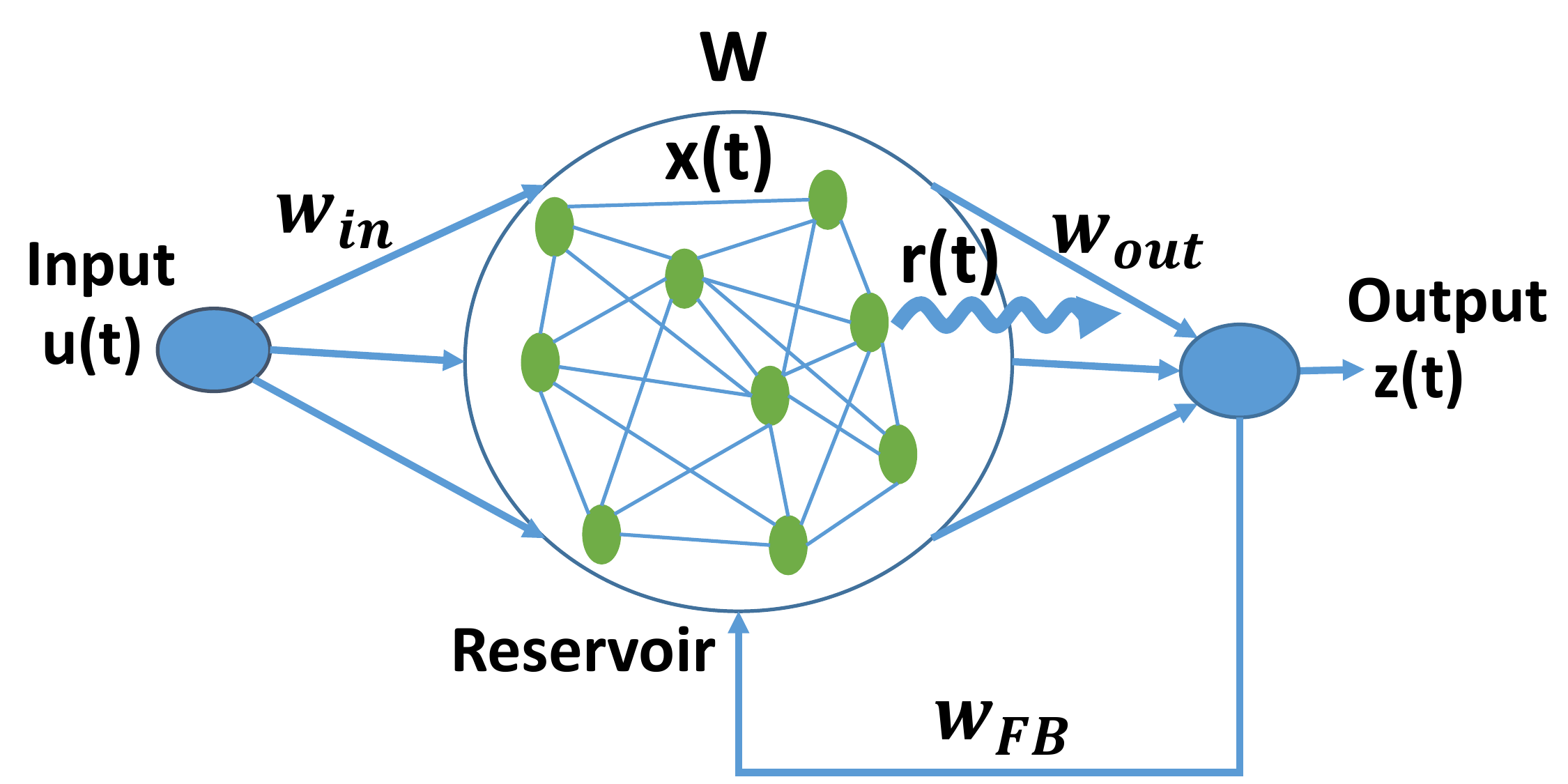}
\caption{Typical Reservoir computing architecture consisting of an input connected to a reservoir of randomly connected neurons. The reservoir activity $r(t)$ is fed into the readout/output neuron, that again feeds back the output activity into the reservoir through weights $w_{FB}$. The remaining notations of the figure are explained and denoted in Eqn. (1).}
\end{figure}

Most of the existing literature tend to bypass `chaos' and theorize the dynamics of reservoir models under designed control settings \cite{sompolinsky1988chaos, rajan2010stimulus, stern2014dynamics, sussillo2013opening} for random conditions. Rivkind et al. \cite{rivkind2017local} demonstrated the first analysis on the effect of training on reservoir dynamics. However, the analysis was restricted to trained models for fixed point targets. In this work, we present a study to understand the dynamics of reservoir models \textit{during training}. As opposed to previous works, we use the `chaotic' activity of the reservoir to gauge and analyze the stability of the model. We consider reservoir networks that have feedback connections from the readout to the reservoir (Fig. 1).  This closed loop setting enables the reservoir to perform complex tasks. But it also poses a major difficulty to analyze the stability of these networks. We break the readout-feedback loop during training by unrolling the network over time (defined as, Breaking the Closed Loop (BCL) methodology). Then, we analyze the activity of the auxiliary open loop system at intermediate time steps at different stages of training. The key contributions of this work are as follows:
\begin{itemize}
    \item We show that the network tends to be less chaotic with training, that is representative of the training success of the network. We evaluate the nonlinear complex dynamics of the reservoir using Eigenvalue (EV) spectra and observe that the spectral circle shrinks as training progresses, representative of decrease in random chaotic projections. 
    \item Furthermore, we extend our analysis to fixed point and time varying targets and observe similar stability behavior. For time varying targets, we use the first order reduced and controlled error (FORCE) training method \cite{sussillo2009generating}, while we use standard least squares regression to train fixed point targets.
    \item Additionally, we use Principal Component Analysis (PCA) to further peek into the model's activity during training. Rajan et al. \cite{rajan2010stimulus, rajan2010inferring} demonstrated that reservoirs operate in a dynamic regime, wherein stable input driven periodic activity and chaotic activity tend to coincide. Interestingly, we observe that our `Breaking the Closed Loop (BCL)' analysis preserves the interaction between the chaotic spontaneous activity and non-chaotic input driven state of the reservoir supporting the results from prior works \cite{rajan2010stimulus, rajan2010inferring, abbott2011interactions}. 
    \item To validate the effectiveness of our BCL methodology, we compared our analysis to the prior work \cite{rivkind2017local}. Rivkind et al. \cite{rivkind2017local} analyze \textit{trained} reservoir dynamics with closed loop theory. We verify that the EV spectra obtained from our proposed BCL after training a reservoir  coincides with the EV spectra obtained from the closed loop theory (under same operating operating conditions).
\end{itemize}
In summary, we analyze the dynamics of a reservoir \textit{during training} and formulate a stability criterion, while substantiating the results from prior work by Rivkind et al.\cite{rivkind2017local} and Rajan et al. \cite{rajan2010stimulus, rajan2010inferring} on trained recurrent dynamics. It is worth mentioning that BCL method of analysing a feedback system can seem as a trivial approach. However, the EV spectra and PCA results obtained from analysing the auxilliary open loop systems yield interesting and novel statistics about the reservoir dynamics and stability, that can be applied to complex dynamical systems for large-scale analysis. 

\section{Model Description and Motivation for BCL}
The dynamics of the reservoir model are given by
\begin{equation}
\frac{dx}{dt} = -x(t) +Wr(t)  + w_{FB}z(t) +w_{in}u(t)
\end{equation}
where $x(t)$ represents the internal state of the reservoir at a given time, $r(t) = \phi(x(t))$ is the neuronal firing rate, where $\phi$ denotes a nonlinear function ($\phi(x) =tanh(x)$ in this work). $z(t) = w_{out} r(t)$ is the output activity of the linear readout neurons that are fed back into the reservoir with feedback weights $w_{FB}$. Input $u(t)$ is fed into the reservoir with input weights $w_{in}$. $W$ represents an $N\times N$ recurrent weight matrix (with $N$ equal to the number of neurons in the reservoir) chosen randomly and independently from a Gaussian distribution with $0$ mean and variance, $g^2/N$, where $g$ is the synaptic gain parameter. $g$ regulates the overall chaotic activity in a system. Previous studies have shown that for large recurrent networks, values of $g >1$ generate increasingly complex and chaotic patterns of spontaneous activity \cite{rajan2006eigenvalue, sompolinsky1988chaos}. In our simulations, we vary $g= 0.9, 1.2, 1.5 $ to understand the behavior of the system under different chaotic conditions. For simplicity in simulations, we follow previous works \cite{rajan2006eigenvalue, rivkind2017local} and set input $u=0$ for all our experiments. In all our simulations, we start by randomly initializing the reservoir state $x(t)$ from Normal distribution and then continue with our training experiments. Note, reservoirs are generally categorised under recurrent neural networks due to the weight matrix $W$ that imparts recurrency to such models \cite{rivkind2017local, sompolinsky1988chaos}. 

The objective of training the readout weights $w_{out}$ is to ensure that the activity of output neurons ($z(t)$) match some predefined target function $f(t)$, i.e. $z(t) = w_{out} r(t) \approx f(t)$. In case of a fixed point target, $f(t) = A$, where $A$ is a constant value. In case of a time varying target, $f(t) = g(t)$, say $g(t) = sin(t)$, where the objective is to train the reservoir to generate sinusoidal activity. It is evident that learning $w_{out}$ for fixed point targets is simple, that involves solving a least squares regression task $w_{out} r(t) = A$ using standard algebraic and linear equation methods. For time varying targets, solving for $w_{out}$ becomes slightly complicated. Thus, we use the popular FORCE training  devised by Sussillo et al. \cite{sussillo2009generating}, a widely used algorithm to train reservoirs for generating sequential patterns. 

A noteworthy observation from Eqn. (1) is that the reservoir model is a closed loop system (due to $w_{FB}$) with perpetual feedback from the output, that affects the internal state $x$ of the reservoir at every time step. Since training occurs in this closed loop dynamics, it is apparent that analyzing the stability of the reservoir will be difficult considering the continuous temporal activity of the system. Here, the feedback affects the training which in turn affects the stability of the system. In several prior works, the authors, thus, analyzed the behavior of random networks without feedback and without any training consideration \cite{sussillo2013opening, rajan2010stimulus, stern2014dynamics}. Such works theorized interesting results pertaining to the stability and robustness criteria for chaotic networks. However, those results cannot be extended to the feedback-based reservoir performing realistic tasks. In fact, Rivkind et al.'s analyses \cite{rivkind2017local} on feedback-based reservoir hinges on the fact that, \textit{the network has already been trained}. Therefore, the reservoir dynamics have converged to a unique stable state. This allows them to use constant feedback that does not affect the reservoir's state. For a fixed point target $f(t) =A$, Rivkind et al. describe the behavior of the trained reservoir model (already at stable state $\bar x$) as
\begin{equation}
\bar x = W \phi(\bar x) + w_{FB} A
\end{equation}
It is clear from Eqn. (2) that the feedback's effect on temporal dynamics is no longer there, since it is constant. Analyzing stability of a reservoir during training requires us to incorporate feedback (that will change with every time step). Hence, we unroll the reservoir dynamics over time without affecting the properties of the system, as described below.

\begin{figure*}[!t]
\centering
\includegraphics[width=0.7\textwidth]{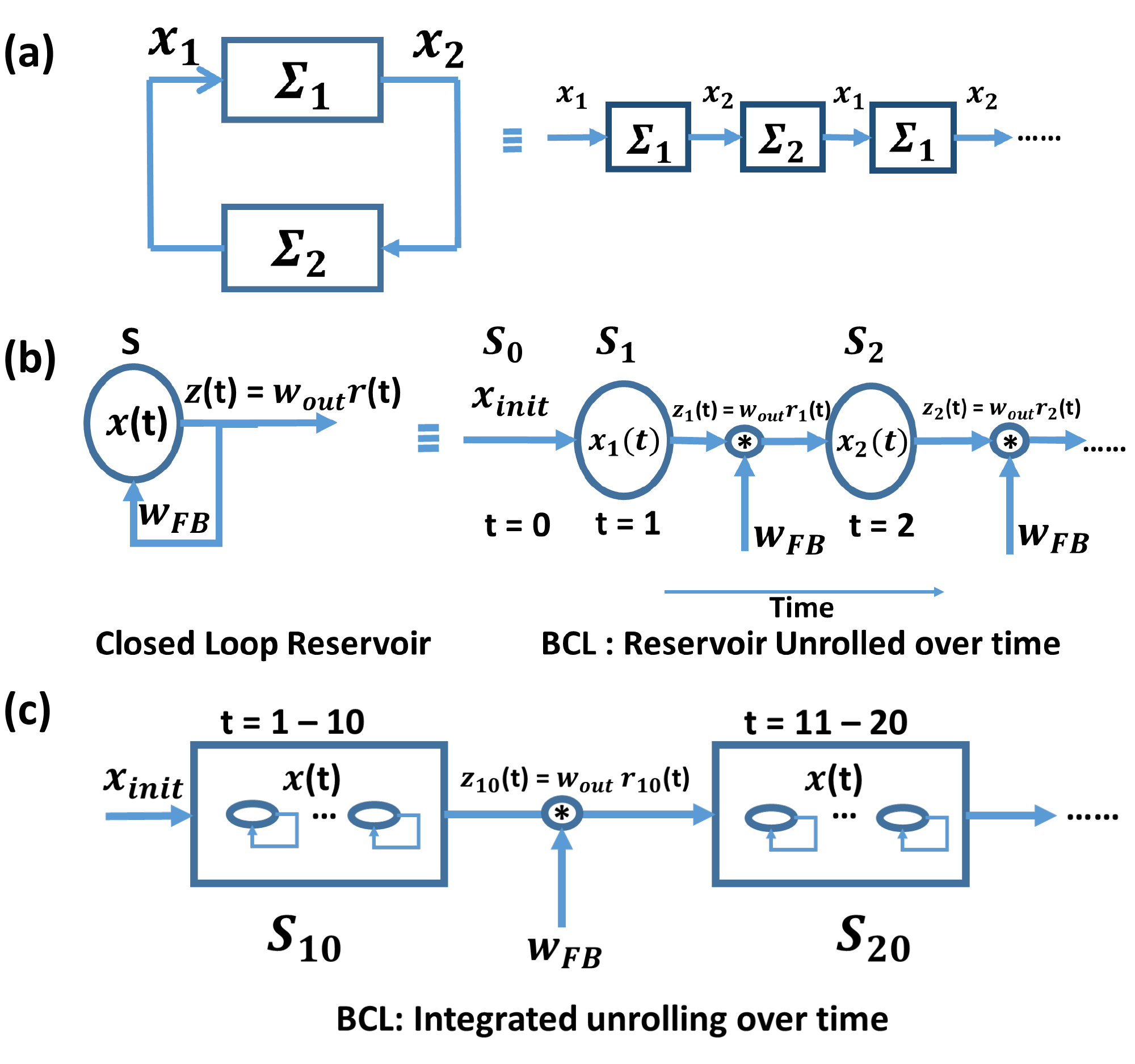}
\caption{(a) Traditional breaking the loop theory used in control systems to analyze a closed loop system as a series of cascaded interconnected open loop systems. (b) The feedback weight in the closed loop reservoir model can be broken using the BCL methodology wherein, the network activity from the previous state is multiplied with $w_{FB}$ to affect the network activity in the current state in the open loop cascaded system. The reservoir dynamics are unrolled over time to preserve the integrity of the system. (c) To avoid unrolling at each time step $dt$ for large scale simulations (that can be expensive), we use BCL while performing integrated unrolling over time. The reservoir dynamics observed at the final time step of the last unrolled instance $z_{10}(t)$ is used with feedback throughout the simulation of the current instance ($t=11-20$).}
\end{figure*}

\section{Breaking the Closed Loop (BCL): Unrolling the Reservoir over Time}
The basic idea of BCL is illustrated in Fig. 2 (a), where the feedback system can be viewed as a series of interconnected cascaded systems unrolled over multiple time steps. As $t\rightarrow \infty$, the cascaded system’s dynamics will converge to that of the closed loop dynamics. This broken loop theory is a widely used concept used in decision and control to analyze the robustness of feedback systems. Here, we take this concept and apply it to analyzing reservoir dynamics during training. Fig. 2(b) shows the unrolled reservoir dynamics over time. Here, the output ($z_i(t)$) from each state ($S_i$) at a given time step $i$ is multiplied by $w_{FB}$ before being fed as input to the network in the next time step. This preserves the temporal property and the feedback dynamics of the reservoir that continually affects the reservoir's internal state $x(t)$ at every time step. Consequently, the reservoir dynamics in the unrolled state can be described as
\begin{equation}
\begin{aligned}
    \frac{dx}{dt} &= -x(t) +Wr(t)  + w_{FB}z_{unroll}(t) \\
    dx &= -x(t) dt +  Wr(t) dt+ w_{FB} w_{out} r(t-1) dt \\
    x(t+1) &= x(t) +dx \\
\end{aligned}
\end{equation}

Eqn. (3) bears resemblance to Eqn. (1). The difference arises from the inclusion of $z_{unroll} (t)$ that accounts for the intrinsic state of the system from the previous time step $r(t-1) = \phi(x(t-1))$ in the unrolled reservoir, to obtain the current state of the reservoir. Note, the input $u(t)$ has not been shown in Fig. 2 and Eqn. (3) for convenience in representation. Now, let us consider training of a network in an unrolled state. The basic idea of training is to learn the weights $w_{out}$ to obtain $z(t) = f(t)$ at each time step. Therefore, $w_{out}$ in the unrolled reservoir system will change over time. For a fixed point target ($f(t)=A)$, this implies solving the linear equation $ w_{out} r_i(t) = A$ for each state $S_i$ before proceeding to calculating the network dynamics $x_{i+1}(t)$ for the next state $S_{i+1}$. It is noteworthy to mention that this simplistic analysis allows us to track the reservoir activity during training without interfering or disrupting the overall dynamics. 

One might argue that unrolling over time will be computationally expensive and time consuming. We need to unroll the system for each time step $dt$ throughout the entire time period of simulation to observe the behavior. However, unrolling time can vary from $dt$. We can possibly integrate the state of the system for several time steps before we unroll. Fig. 2 (c) shows the system unrolled at every $10th$ time step. In this case, the state of the system for the next $t=11 - 20$ integration will utilize the last reservoir state, i.e. $S_{10} \equiv r_{10}(t)$, as the unrolled feedback input $z_{unroll}(t)$. This approximation helps us circumvent the computational issue without affecting the overall dynamics of the system, while allowing us to perform the stability analysis for each unrolled reservoir state. Note, for integrated unrolling, the last reservoir state after every unrolling (for instance, $S_{10}, S_{20} …$ in Fig. 2 (c)) is used to gauge the stability of the system.  Next, we describe the eigenvalue (EV) spectra method that measures the stability of the unrolled reservoir dynamics.

\section{Eigenvalue (EV) Spectra: Stability Evaluation Criteria}
EV spectra is a powerful tool of random matrix theory that allows one to examine the complex behavior of reservoir networks with random recurrent connections. By diagonalizing the synaptic weight matrix $W$ of the reservoir, we obtain complex modes that represent the activity (specifically, the frequency of oscillation) of each neuron in the reservoir \cite{rajan2006eigenvalue, rajan2009spontaneous}. Each mode is denoted by a complex eigenvalue, where the real part ($Re(EV)$) denotes the decay rate of the neuronal oscillation and the imaginary part ($Im(EV)$) denotes the frequency of oscillation. It has been shown in several works that a neuron or mode with $Re(EV) > 1$ exhibits long-lasting oscillatory behavior representative of \textit{chaos} \cite{rajan2006eigenvalue, rajan2009spontaneous}. The authors in \cite{panda2018learning, panda2017learning,rajan2009spontaneous} have shown that a reservoir model with good ‘memory’ must operate in a region between singular fixed point activity ($Re(EV) << 1$) and complete chaos ($Re(EV) >> 1$). In the context of learning, the reservoir's activity which is generally chaotic in the beginning of training must converge to stable states as training proceeds. That is, there should be fewer modes with $Re(EV) > 1$ as training progresses. This suggests that the EV spectral circle must shrink to ensure the success of training. In fact, recent works that use novel plasticity rules to train the recurrent weights of the reservoir have demonstrated the effectiveness of their learning methodology with EV spectral circle shrinking \cite{panda2018learning, panda2017learning}.

We take this EV spectra evaluation criterion and apply it to analyze the activity of the reservoir at different unrolled instances. Linearizing the reservoir's dynamics will model the diagonalization of synaptic weights $W$, which in turn, determines the EV spectra. Linearizing Eqn. (1) which is a closed loop system gives
\begin{equation}
\delta \dot x = [-I +W*r' + w_{FB}w_{out}r']\delta x
\end{equation}

In fact, Rivkind et al. analyzed the EV spectra of the trained reservoir using Eqn. 4. In our case, as mentioned earlier, we analyse the network activity after every unrolling to gauge the stability of the system. Linearizing the unrolling dynamics (Eqn. (3)) yields
\begin{equation}
\delta \dot x = [-I +W*r' + w_{FB}w_{out}r'_{unroll}]\delta x
\end{equation}
We would like to emphasize that $r_{unroll}$ is the reservoir activity from the previous time step or unrolling instance. In case of integrated unrolling (refer Fig. 2 (c)), the EV spectra of the reservoir state $S_{20}$ after the second unrolling at $t=20$ uses $r_{unroll} =r_{10}$ value from last time step of the previous integrated unrolled instance. Note, in our simulations done in MATLAB, we use the available $eig$ tool to plot the EV spectra. This translates Eqn. (5) to $-1 + eig(Wr’ + w_{FB} w_{out} r’_{unroll})$ in the simulation framework (note, proper usage of $transpose$ and $diag$ functions (not shown here) are necessary to maintain the dimensionality of the matrices during implementation). 

\section{Results}
\subsection{EV spectra to evaluate training stability}
We conducted reservoir training with the unrolling BCL methodology and gauged the stability of training by evaluating the EV spectra. First, we trained a reservoir of 1000 neurons to generate a fixed point target ($f(t) = A =1.5$) with least squares regression training for varying $g$ values. Since the fixed point target is a simple target, the reservoir could easily get trained in $t=100-250$ timesteps with $dt=1$. As a result, we did not have to use integrated unrolling. The results are shown in Fig. 3 for $g =0.9, 1.2, 1.5$. Note, the accuracy of the reservoir undergoing BCL unrolling is same as that of performing training on a Closed Loop (CL) reservoir system for all experiments in Fig. 3. 

\begin{figure*}[!t]
\centering
\includegraphics[width=\textwidth]{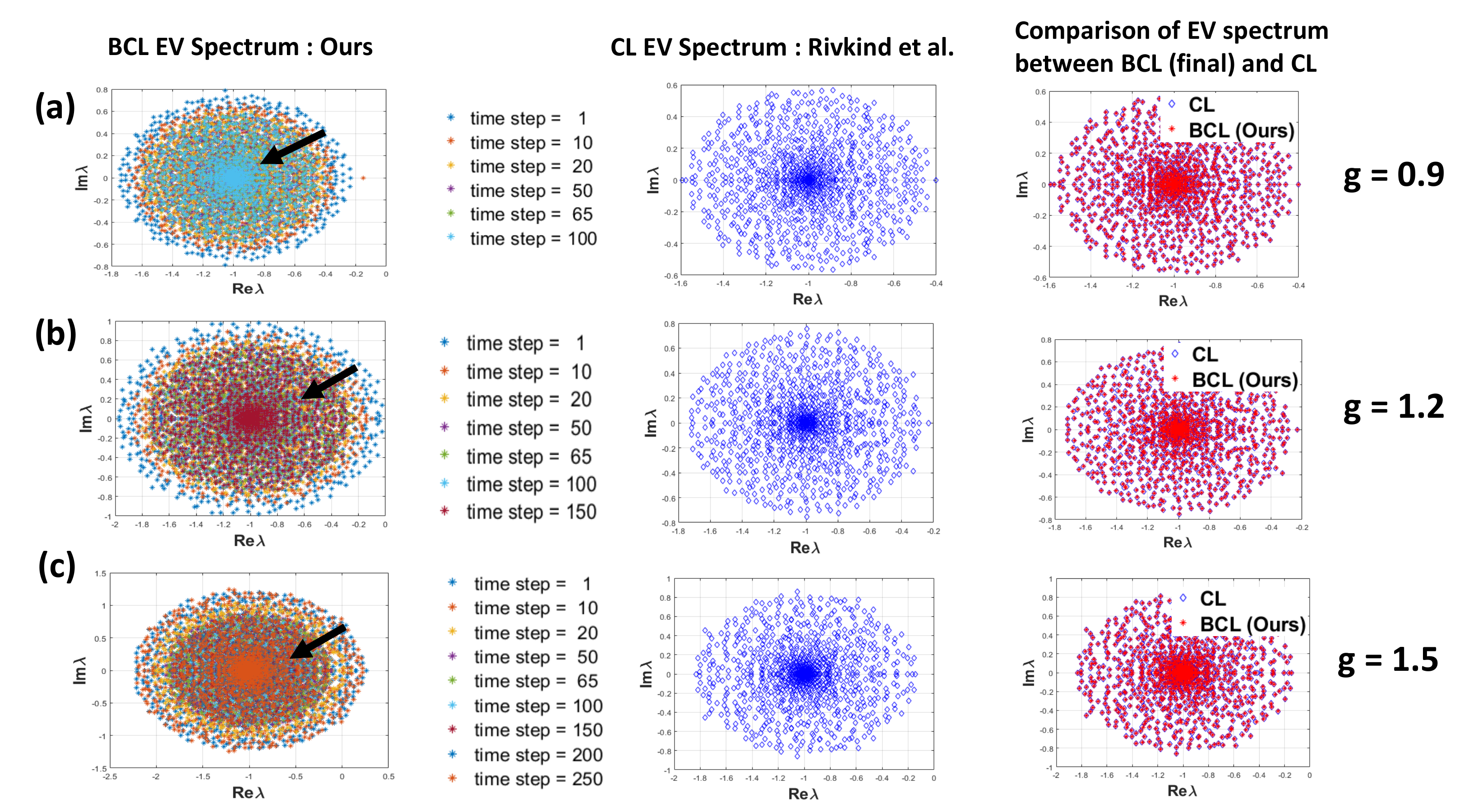}
\caption{EV spectra for varying values of \textit{g = a) 0.9, b) 1.2, c) 1.5}. Column 1 shows the EV spectra evaluated with our methodology at various time steps during training. The black arrow represents the shrinking of the spectra with increasing time steps signifying stability and convergence behavior with reduced chaotic activity. Column 2 shows the EV spectra plotted on a trained reservoir model using the Closed Loop (CL) methodology proposed by Rivkind et al. Column 3 validates the effectiveness of our BCL methodology showing that the EV spectrum obtained from our BCL method at the final time step coincides with that of Rivkind et al.}
\end{figure*}

Fig. 3 (Column 1) illustrates the EV spectra measured from the reservoir dynamics at different steps. It is evident that the spectral radius decreases with increasing time implying the success of the training algorithm in converging the reservoir's chaotic state to stable fixed point activity. To quantify this further, we measured the radius of the spectrum at the initial and final time step for each case. Table 1 illustrates the radii results that further demonstrates shrinking of EV spectra during training. In Fig. 3 (Column 2), we plot the EV spectra of the reservoir system using Rivkind et al.'s Closed Loop (CL) dynamics (Eqn. (2)). We use the open source code available from the authors to perform this analysis. We use the same operating conditions across ours and their method for iso-comparison. In Fig. 3 (Column 3), we plot the EV spectrum obtained at the final time step or unrolling instance from our BCL methodology and compare with that of the spectrum of Rivkind et al.'s CL method. The EV spectra coincide validating our methodology and stability analysis. Rivkind et al. use the EV spectrum to show that the dynamics of the trained reservoir is in a stable regime after training. We get the same spectrum with BCL unrolling at the final step (with a slowly evolving spectra in the intermediate time steps). This establishes the effectiveness of BCL to gauge the stability of a reservoir undergoing training. 

Furthermore, the coincidence of the final timestep EV spectrum from our analysis with that of Rivkind et al. also implies that: \textit{our stability analysis at each unrolling timestep of an open loop system provides a rigorous assessment of the whole closed loop feedback system}. We observe similar behavior across all $g$ values. An interesting observation here is that the total time for convergence increases with increasing $g$ (for instance, $t=100$ for $g=0.9$ to $t=250$ for $g=1.5$). This is expected as $g$ determines chaotic activity. Thus, a reservoir with abundant chaotic projections will take more time to converge during training. Please note, due to the numerical nature of the simulations, we imposed a stopping criteria where we put an upper bound on the maximum time period of convergence($t\le800$ time steps) OR an upper bound on the weight difference between consecutive time steps ($|w_{out}(t)-w_{out}(t-1)| \le 1e-5$). 

\begin{table}
\label{table1}
\centering
{\normalsize
\begin{tabular}{|c|c|c|c|}
\hline
Target & g & $t_{initial}$ & $t_{final}$\\
\hline
\multirow{3}{6em}{Fixed Point}& 0.9 & 0.779 & 0.587 \\
& 1.2 & 0.963 & 0.708 \\
& 1.5 & 1.176 & 0.814 \\
\hline
Time-Varying & 1.5 & 1.266 & 1.036 \\
\hline
\end{tabular}}
\caption{EV spectra radius measured at initial and final time step for different experiments: Fixed point target in Fig. 3 (Column 1: a, b, c for different $g$ values), Time-varying target in Fig. 4 (a) for $g$ =1.5. Note $t_{initial}=1$ is the initial time step in all experiments. $t_{final} = 100, 150, 250$ for Fig. 3 (a) or g=0.9, Fig. 3 (b) or g=1.2, Fig. 3 (c) or g=1.5 respectively. $t_{final} = 5600$ for Fig. 4 (a) corresponding to time-varying target experiment.}
\end{table}

\begin{figure*}[!t]
\centering
\includegraphics[width=0.7\textwidth]{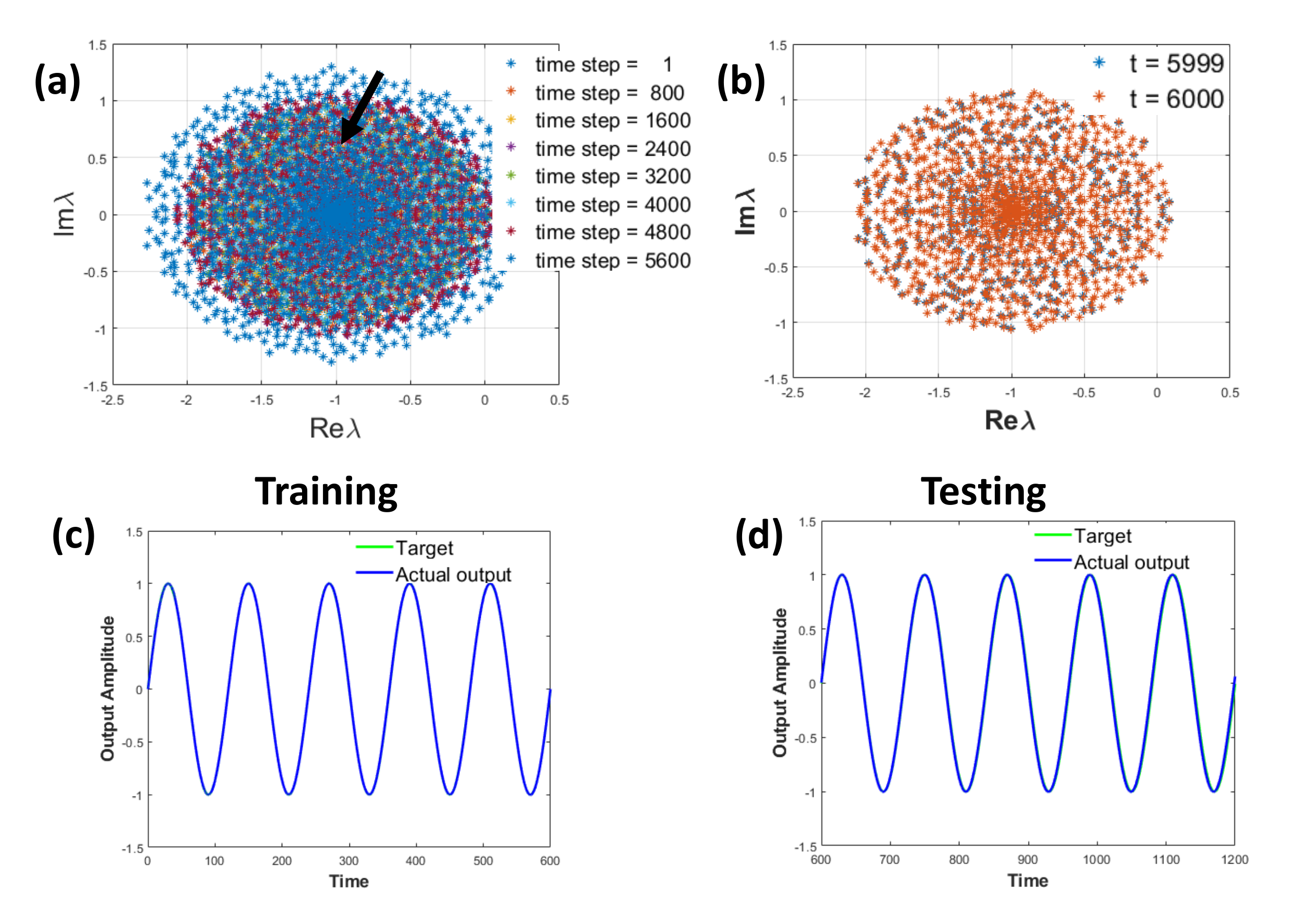}
\caption{EV spectra observed during training of a reservoir model to generate a sinusoidal target. Here, the unrolling was perfomed at each time step $dt=1$. (a) EV spectra converges with increasing training time showing reduced chaos and stability behavior. (b) EV spectrum at the final time steps coincide establishing the convergence of training behavior. The output activity $z(t)$ and target curves obtained during (c) training and (d) testing. We observe a complete match between target and output showing the success of the training algorithm.}
\end{figure*}

\begin{figure*}[!t]
\centering
\includegraphics[width=\textwidth]{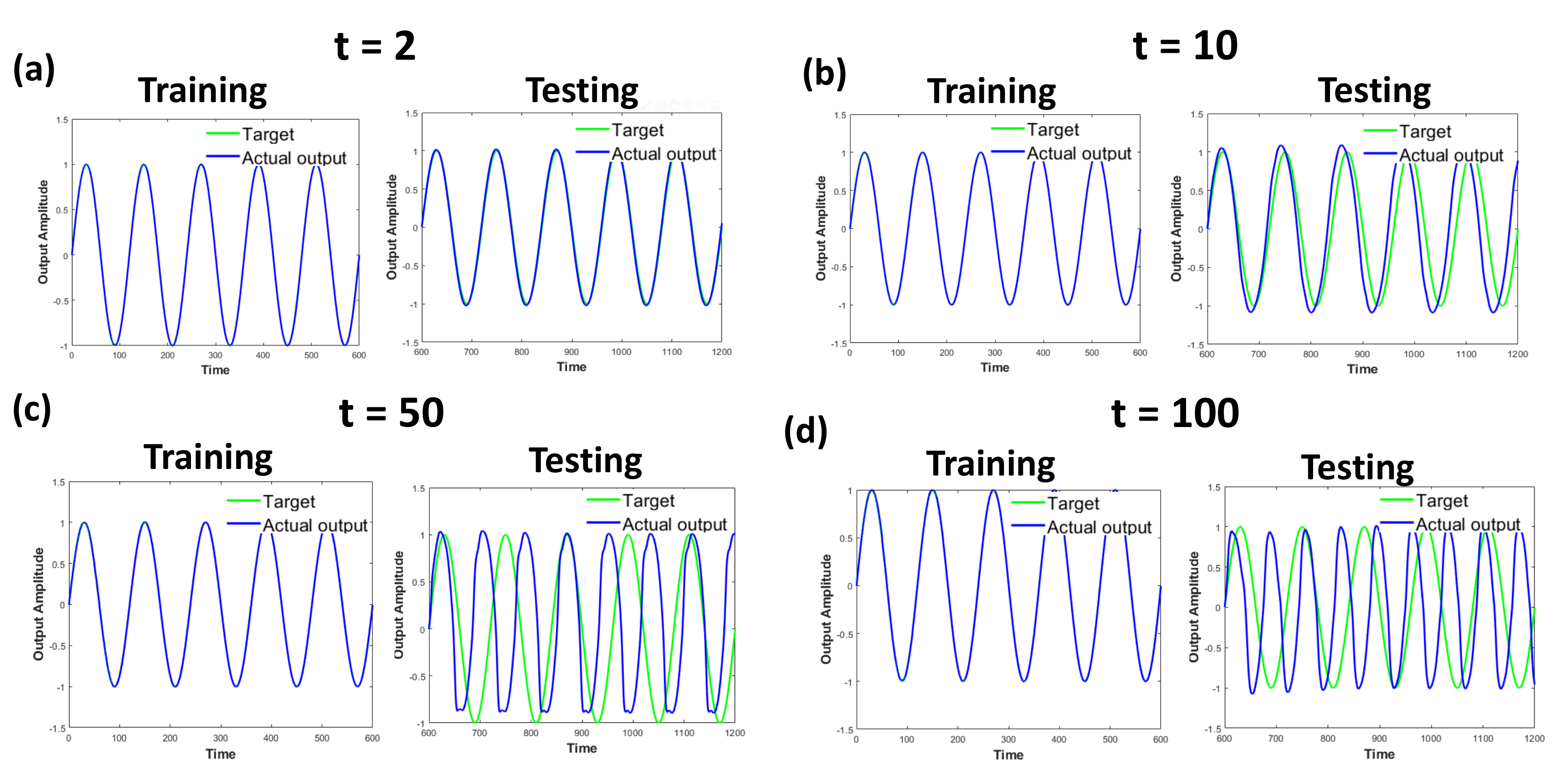}
\caption{The output activity and the target are shown for different training and testing instances when we perform integrated unrolling over time. The unrolling is performed at interleaved time steps with time intervals corresponding to (a) $t=2$ (b) $t=10$ (c) $t=50$ (d) $t=100$. Training is not affected due to unrolling. However, the testing curves show significant degradation in performance as the time interval between two unrolling instances increases.}
\end{figure*}

\begin{figure*}[!t]
\centering
\includegraphics[width=0.65\textwidth]{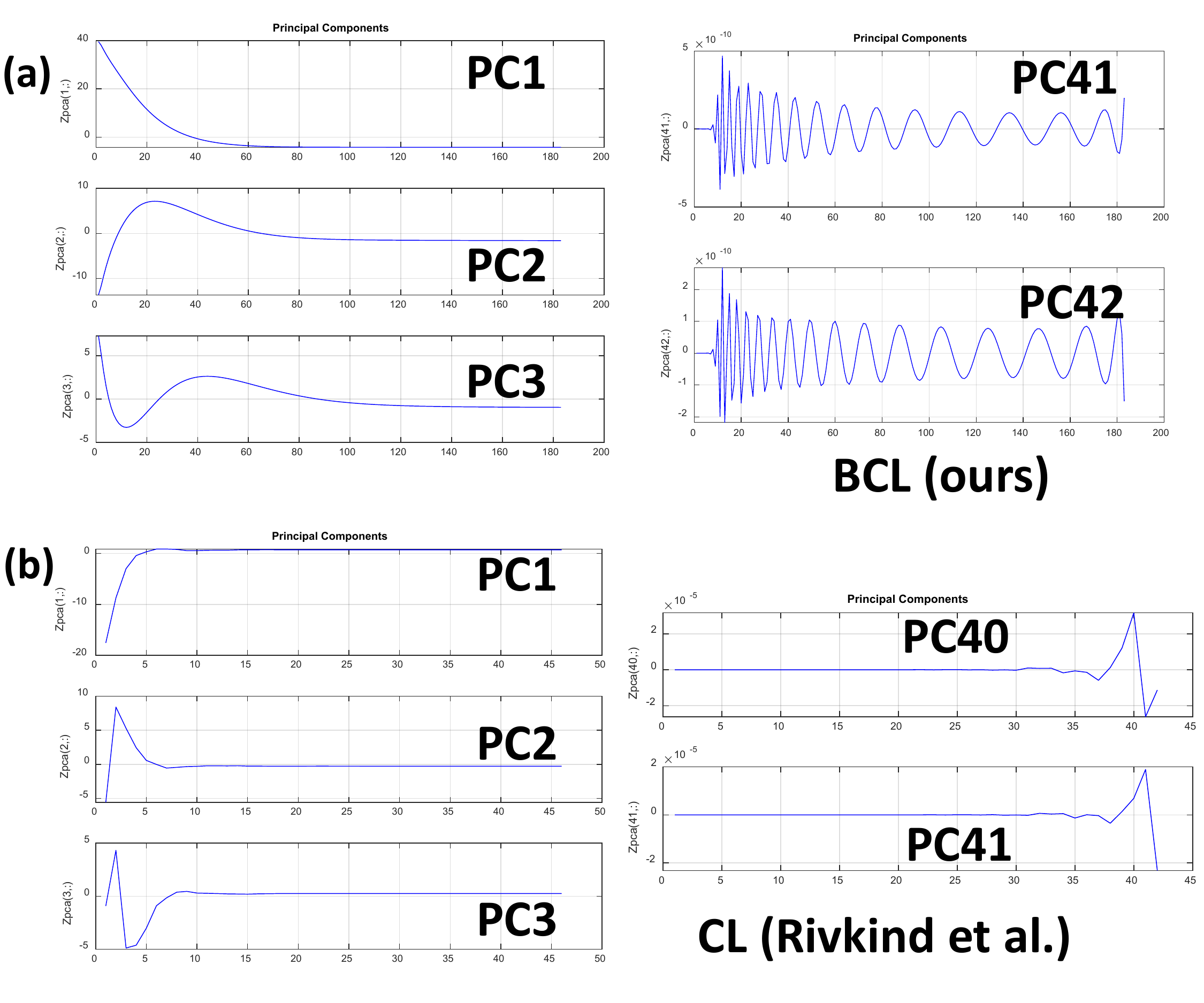}
\caption{Trajectories corresponding to various Principal Components of a trained reservoir model is shown. (a) We obtain the PCs after training a reservoir for fixed-point target (similar to Fig. 3(a)) with our BCL unrolling over time methodology. (b) We obtain the PCs of a trained network using the closed loop methodology proposed by Rivkind et al.}
\end{figure*}

Next, we analyzed a reservoir being trained for time varying targets to generate a sinusoidal ($f(t) = Sin(20\pi t)$) pattern. We simulated a reservoir of 1000 neurons (with $g=1.5$) with FORCE training. Fig. 4 shows the results for EV spectra observed at several time steps during training. It is clear that the spectrum shrinks over time (also quantified by the radii results in Table 1). In fact, the spectrum at the final time steps: $t=5999, 6000$, completely match signifying the convergence of the training algorithm. We also see that the output activity during training and testing matches with that of the target, empirically indicating the success of the training rule. In Fig. 4 analysis, the network was unrolled for each time step $dt=1$. As a result, the simulation time was longer ($\sim$2 minutes as per real clock time). In contrast, using integrated unrolling over time (as discussed in Section 3) decreases the overall simulation clock time at the cost of degraded convergence. 

Fig. 5 compares the output activity and target for varying integrated unrolling time instances ($t = 2, 10, 50, 100$). We see that as the unrolling takes place at increased interleaved time steps, the output activity fails to match the target activity during testing. While the curves match during training, the testing fails with the output activity significantly shifting away from the target as the integrated time between unrolling increases. The shift during testing is indicative of a decline in accuracy. This is expected as the unrolling behavior approximates the network dynamics. And, with interleaved unrolling, certain significant aspects of the temporal dynamics might get affected. Please note, the EV spectra for the reservoir dynamics corresponding to Fig. 5 (not shown), specifically for $t= 50, 100$ do not exhibit shrinking behavior that further corroborates the empirical results seen in Fig. 5. We, therefore, recommend using unrolling at each time step to attain reliable results.

We would like to emphasize that the presented analysis is the first work to show the stability of training a recurrent model with time varying target. We would also like to note that our experiments were restricted to simple fixed-point or time-varying target analysis due to the intensive numerical nature of simulations as well as the limitation of reservoir training methods. We believe that our analysis can be extended to complex problems on real-world datasets, given that the reservoir training in such cases can be done effectively.

\subsection{PCA to analyse BCL methodology}
Rajan et al. \cite{rajan2010inferring} demonstrated that reservoirs (that have converged to a given state) exhibit chaotic as well as stable periodic activity. They use PCA to analyze the network activity and visualize the stable and chaotic trajectories. The network state at any given time instant can be described by a point in the \textit{N}-dimensional space with coordinates corresponding to the firing rates of the \textit{N} neuronal units. With time, the network activity traverses a trajectory in this N-dimensional space and PCA can be used to visualize the trajectory. To conduct PCA, we diagonalize the equal-time cross-correlation matrix of the firing rates of the \textit{N} units as
\begin{equation}
D_{ij} = <(r_i(t)-<r_i>)(r_j(t)-<r_j>)> 
\end{equation}
where the angle brackets, $<>$, denote time average. The eigenvalues of the matrix $D$ (specifically, $\lambda_a/\sum_{i=1}^{N}\lambda_a$, where $\lambda_a$ is the eigenvalue corresponding to principal component $a$) indicate the contribution of different Principal Components (PCs) toward the fluctuations/total variance in the spontaneous activity of the network. Rajan et al. observed that the network activity shows fluctuating patterns and irregular trajectory in the higher order PCs (such as $PC>10$) characteristic of chaos. In contrast, the trajectories for lower order PCs are more regular and non-fluctuating characteristic of stability. To further corroborate the effectiveness of BCL for analyzing the training stability of reservoir models, we plotted the PCs for different components (PC 1, 2, 3, 41, 42) in Fig. 6 for a model trained with fixed point target (corresponding to Fig. 5 (a)). Note, the network activity was observed after training using the BCL unrolling methodology. We observe slowly fluctuating patterns for PC 1, 2, 3, while extensively fluctuating patterns for PC 41, 42. 

Plotting the PC curves for the reservoir trained with CL dynamics as proposed by Rivkind et al., we observe non-fluctuating pattern of activity across both high and lower order PCs. This suggests an absence of chaotic activity in the system. However, as Rajan et al. have demonstrated, there will always be chaos in a reservoir coinciding with stable patterns of activity. We believe that the assumption Rivkind et al. make regarding the convergence of the system to stable state before analyzing the network activity (Eqn. (2)) causes such discrepancy.  PCA results further establish the correctness of our methodology in preserving the integrity of the system and its temporal dynamics.

\section{Conclusion}
We present a first of its kind methodology to analyze the stability of reservoir models (with feedback) during training. Essentially, we unroll the reservoir dynamics over time and analyze the eigenvalue spectra of the reservoir. The shrinking spectra during training underscores the success of the training methodology while signifying the convergence of the reservoir's chaotic activity to more convergent stable states. To minimize the number of unrolling time steps for large scale simulations, we also presented the integrated unrolling over time methodology. However, we observed that the accuracy of the system gets affected when the unrolling instances are done over longer time intervals. We showed the effectiveness of our proposed methodology for training reservoir models on fixed point as well as time varying targets. Our analysis establishes \textit{eigenvalue spectra} /\textit{breaking the closed loop} methodology as a reliable metric/technique to evaluate/gauge the stability of training in reservoir models, respectively. In the future, we would like to extend this proposal to analyze the robustness of the network prediction and examine the interpretability of the network's behavior during and after training.

\section*{Acknowledgement}
The work was supported in part by, Center for Brain-inspired Computing Enabling Autonomous Intelligence (C-BRIC), a DARPA sponsored JUMP center, by the SRC, the NSF, Intel Corporation, the DoD Vannevar Bush Fellowship and by the U.S. Army Research Laboratory and the U.K. Ministry of Defense under Agreement Number W911NF-16-3-0001.


\end{document}